\documentclass[runningheads]{llncs}
\usepackage[numbers]{natbib}
\usepackage{times}
\usepackage{latexsym}
\usepackage{multirow}
\usepackage{booktabs}
\usepackage{graphicx}
\usepackage{comment}
\usepackage{soul} 

\PassOptionsToPackage{hyphens}{url}\usepackage{hyperref}

\usepackage[textsize=scriptsize,backgroundcolor=green]{todonotes}

\begin{document}
\title{The Case for Claim Difficulty Assessment in Automatic Fact Checking} 

\author{Prakhar Singh \and Anubrata Das \and Junyi Jessy Li\thanks{Both authors contributed equally.} \and Matthew Lease$^\star$}
\institute{University of Texas at Austin\\
\email{prakharsingh95@gmail.com\\\{anubrata@, jessy@austin., ml@\}utexas.edu}}

\maketitle

\begin{abstract}
Fact-checking is the process (human, automated, or hybrid) by which {\em claims} (i.e., purported facts) are evaluated for veracity. In this article, we raise an issue that has received little attention in prior work -- that some claims are far more difficult to fact-check than others. We discuss the implications this has for both practical fact-checking and research on automated fact-checking, including task formulation and dataset design. We report a manual analysis undertaken to explore factors underlying varying claim difficulty and categorize several distinct types of difficulty. We argue that prediction of claim difficulty is a missing component of today's automated fact checking architectures, and we describe how this difficulty prediction task might be split into a set of distinct subtasks. 
%
\end{abstract}



\section{Introduction}

Recent years have seen a surge of misinformation (and closely related phenomena, such as disinformation and fake news), and that such information can cause serious societal harm. For example, such detrimental effects include (among others) influencing voting decisions in an election, health care decisions during a pandemic, or stock prices\footnote{e.g., the GameStop incident in 2021: \url{https://reut.rs/3ra7stz }}. 

The rise of misinformation has also prompted a great body of work, especially in natural language processing (NLP), on the automatic fact checking of claims
\cite{fever,popat2018declare,multifc,hanselowski2019richly,wadden2020fact,xfact,covidfact}. Despite tremendous progress, however, the task remains quite challenging. For example, recent performance on key datasets for natural claims is less than 50\% F1~\cite{Atanasova2020GeneratingFC,multifc,xfact}
Digging further into the data reveals that some claims are far more difficult to fact-check than others, an issue that has received little attention. 

{\bf Table~\ref{tab:taxonomy-examples}} shows several claims that were fact-checked by Politifact\footnote{\url{https://www.politifact.com}}. If we compare claims 3 and 5, claim 3 can be directly verified by a simple search leading to a major media news article. On the other hand, for claim 5 there does not appear to be any direct sources (i.e. written evidence, online or otherwise). In fact, when historians were eventually consulted, even they asserted that this claim was ultimately uncertain.
These examples illustrate a vast spectrum of varying difficulty in checking different claims.

In this work, we present a first investigation into why some claims are more difficult to fact-check than others, based on  manual analysis of claims from  Politifact\footnote{\url{https://www.politifact.com/}} that have been used in automated fact-checking datasets \cite{liar,liarplus,multifc}. This analysis leads us to identify the following five distinct factors contributing to claim difficulty: claim ambiguity, source support, source readability, source trustworthiness, and inference. 

Incorporating claim difficulty into fact-checking has conceptual, practical, and research implications. Conceptually, while prior work has considered whether a given claim is check-worthy \cite{checkthat2020,nakov2021clef}, a claim may be check-worthy yet be extremely difficult or impossible to fact-check. This has consequences both for practical fact-checking, where human resources are limited and must be thoughtfully prioritized, and for dataset design for automated fact-checking, where inclusion of extremely difficult claims may provide no useful insights for comparing alternative algorithms. Moreover, automated prediction of claim difficulty could be useful to fact-checking organizations in resource prioritization, yet we are not familiar with any research on this prediction task.

While claim difficulty prediction appears novel in the context of fact-checking, we note that the value of modeling and predicting difficulty of different task instances is already recognized in other areas, such as machine translation  \cite{Mishra2013AutomaticallyPS,Li2015DetectingCS}, syntactic parsing \cite{Garrette2013LearningAP}, or search engine switching behaviors \cite{white2009characterizing}, among others. From this perspective, we argue instance difficulty prediction can bring similar value to fact-checking.

\section{Claim Difficulty Analysis}
\label{sec:claim-difficulty-analysis}

We conduct our manual analysis over several NLP fact-checking datasets that use Politifact data~\cite{liar,liarplus,multifc,fakenewsnet}, noting that the accuracy of fact checking models has tended to be lower on claims sourced from Politifact vs.\ other sources. Whereas \citet{Vlachos2014FactCT} focus on expert sources, we consider issues in the absence of expert sources.




\begin{table*}[ht!]
\centering
\resizebox{\textwidth}{!}{%
\begin{tabular}{@{}cp{12.5cm}cc@{}}
\toprule
\#         & \textbf{Example}                                                                                                                                                                                                                                                                                                                                                                                                                                                 & \textbf{Verdict}                     & \textbf{Difficulty}                                                                           \\ \midrule
\textbf{1} & \textbf{Midtown crime is up by 30 percent the last quarter. {[}Crime, Florida, Rick Baker{]}}\\ &\footnotesize{\path{https://www.politifact.com/factchecks/2017/jul/11/rick-baker/did-crime-go-30-percent-midtown-last-quarter/}}                                                                                                                                                                                                                                                                                                                                               & \multirow{4}{1cm}{Mostly False}  & \multirow{4}{2cm}{\textbf{Unjudgable}. Claim is ambiguous. Which midtown?}                                   \\
\textit{a} & \textit{{[}therealdeal.com{]} Manhattan leasing strong in Q3; Midtown market stratifies ...}                                                                                                                                                                                                                                                                                                                                             &                                                                                                                    \\
\textit{b} & \textit{{[}www.news-leader.com{]} Springfield police say a spike in stolen cars led to a 2 percent increase in crime in the third quarter of 2016}                                                                                                                                                                                                                                                                                       &                                                                                                                   \\
\hline
\textbf{2} & \textbf{There are more members of the U.S. Senate than the number of WI families who would benefit from GOP estate tax break. {[}Taxes, Wisconsin, Tammy Baldwin{]}} \\
& \footnotesize{\path{https://www.politifact.com/factchecks/2015/may/20/tammy-baldwin/are-there-more-us-senators-wisconsin-residents-who/}}                                                                                                                                                                                                                                                                     & \multirow{4}{1cm}{Mostly True}   & \multirow{4}{2cm}{\textbf{Hard}. Poor source support. Lots of search results but none relevant.}       \\
\textit{a} & \textit{{[}politicalticker.blogs.cnn.com{]} The candidates for Senate in Wisconsin stuck to their talking points in an animated debate Monday night...}                                                                                                                                                                                                                                                                                  &                                                                                                                   \\
\textit{b} & \textit{{[}www.dailypress.com{]} The Senate ... voted for a new tax break benefiting companies that shifted headquarters to foreign countries, but move back and invest in ...}                                                                                                                                                                                                                                                          &                                                                                                               \\
\hline
\textbf{3} & \textbf{Mark Pocan’s proposal to eliminate the U.S. Immigration and Customs Enforcement (ICE) would "eliminate border enforcement. [Immigration, Legal issues, Crime, Regulation, Wisconsin, Leah vukmir]}\\ & \footnotesize{\path{https://www.politifact.com/factchecks/2018/jul/13/leah-vukmir/gop-us-senate-hopeful-leah-vukmirs-pants-fire-clai/}}                                                                                                                                                                                                                                                                                                              & \multirow{4}{1cm}{Pants on Fire} & \multirow{4}{2cm}{\textbf{Easy}. Well supported by trustworthy sources. Claim directly validated.}     \\
a          & \textit{{[}www.vox.com{]} And it was split into three separate agencies: Legal immigration processing went to US Citizenship and Immigration Services, border enforcement went to Customs and Border Protection, and interior enforcement went to ICE.}                                                                                                                                                                                  &                                                                                                                \\
b          & \textit{{[}news.yahoo.com{]} Haggman said his reply is to point out that ICE is an internal enforcement agency and its absence wouldn’t affect border security.}                                                                                                                                                                                                                                                                         &                                                                                                                \\
\hline
\textbf{4} & \textbf{Premeditation, in murder cases like the Oscar Pistorius case, can be formed in the twinkling of an eye. [Criminal justice, Legal issues, Crime, Nancy grace]} \\
& \footnotesize{\path{https://www.politifact.com/factchecks/2014/mar/31/nancy-grace/nancy-grace-talks-oscar-pistorious-trial-could-pre/}}                                                                                                                                                                                                                                                                                                                              & \multirow{5}{1cm}{Mostly False}  & \multirow{5}{2cm}{\textbf{Hard}. Well supported but all sources are not trustworthy.}                  \\
a          & \textit{{[}Fox News Transcript{]} in law school they teach premeditation can be formed in an ininstant. people think you have to plan it out in advance. the reality is here this was an emotional crime.}                                                                                                                                                                                                                               &                                                                                                               \\
b          & \textit{{[}User comment{]} In other words, the killer had to have formed the present intention of killing her while not in the heat of passion or rage, and carried out the killing with the specific intention of killing her. Nateyes is right that it can happen in a matter of minutes. at least under US law, but there has to be some time to reflect and the very specific state of mind of planning and intending to kill.}      &                                                                                                            \\
c*         & \textit{{[}dutoitattorneys.com{]} In a recent High Court ruling in the matter of State v Raath, the full bench of the Cape High Court found that where a father forced his son to remove a firearm from the safe so that he could kill the son’s mother was not sufficient to constitute premeditated murder and it is with that in mind that we are of the view that there is no premeditation at this stage present in Ocar Pistorius’ matter.} &                                                                                                        \\
\hline
\textbf{5} & \textbf{When President Abraham Lincoln signed his Emancipation Proclamation, ``there were over 300,000 slaveholders who were fighting in the Union army.'' [History, Chuck Baldwin]}\\
& \footnotesize{\path{https://www.politifact.com/factchecks/2015/jul/27/chuck-baldwin/no-300000-slave-owners-did-not-fight-union-side-ci/}}                                                    
& \multirow{4}{1cm}{Pants on Fire} & \multirow{4}{2cm}{\textbf{Unjudgable}. Well supported by trustworthy sources but claim cannot be validated.} \\
a          & \textit{{[}American Studies Journal{]} Eventually 184,000 black soldiers fought for the Union army and contributed significantly to the victory of the North. Reports of atrocities among black Union soldiers had shocked Lincoln, and he subsequently issued the General Order No. 252 on July 30, 1863 as an Order of Retaliation that clearly reflected his stance towards the status of African American soldiers}                  &                                                                                                                \\
b          & \textit{{[}International Studies Review{]} {[}T{]}he number of actual slaveholders in the South of the Union does not amount to more than 300,000, a narrow oligarchy that is confronted with many millions of so-called poor whites...}                                                                                                                                                                                                 &                                                                
\\
\bottomrule
\end{tabular}
}
\vspace{0.5em}
\caption{ \label{tab:taxonomy-examples} Examples illustrating various aspects of claim difficulty. Claims are in \textbf{boldface}, metadata is in [brackets] and snippets from Bing search results are \textit{italicized}. * denotes source cited by Politifact but not found by Bing. We submitted claims verbatim to Bing to find up to 50 search results with the constraint that search results predate Politifact's publication to simulate a realtime search. We only show the most relevant search results for brevity.}
\end{table*}

\subsection{Claim ambiguity}\label{sec:ambiguity}

Fact-checking a claim requires understanding it well enough to know what should be checked. Ambiguity is thus a natural obstacle. While \citet{Vlachos2014FactCT} entirely exclude ambiguous claims, this may be too restrictive. Natural language use is rife with ambiguity and systems today can and do resolve many forms of it.  As such, we might instead consider in dataset design what sorts of ambiguity fact checking systems are likely to encounter in practice and might be reasonably expected to resolve.

For example, in the first claim in Table~\ref{tab:taxonomy-examples}, does the entity ``Midtown'' refer to Midtown NYC, Midtown WI, or some other Midtown? Similarly, the relative temporal expression ``last quarter'' also requires temporally grounding the claim. In general, claim ambiguity can arise in a multitude of ways, being lexical, syntactic, semantic, or pragmatic.  It may arise due to unresolved entities, ambiguous pronouns \cite{Kocijan2020ARO}, or  ill-defined terms \cite{Vlachos2014FactCT}. In some cases, other text within the claim may enable disambiguation; for example, with the second claim in Table~\ref{tab:taxonomy-examples}, the abbreviation ``WI'' could be reasonably resolved to be the state of Wisconsin given ``U.S. Senate'' in the claim text. However, in other cases, additional context beyond the claim text may be required, such as the date of a given claim. Ultimately, if we cannot resolve ambiguity, we must check all options or rule that the claim cannot be checked because it is unclear what has been claimed.

\citet{fever} resolve named entities in their dataset to reduce ambiguity, while other datasets provide contextual metadata \cite{liar,multifc}. For instance, the Politifact tags for claim 2 contain ``Wisconsin'' which clearly disambiguates ``WI''. Such ambiguity should also be taken into account during dataset construction when search engines are used to obtain relevant evidence to support automated claim checking experimentation. Naturally, search based on ambiguous claims can yield poor quality search results, and thus insufficient evidence if included in a dataset to facilitate research on evidence-based fact-checking. To the best of our knowledge, this has not been considered with most existing datasets~\cite{liar,liarplus,Popat2016CredibilityAO,fakenewsnet,multifc}, though recent work on fact checking related to COVID-19 did usefully evaluate pipeline systems using Google as a baseline engine~\cite{covidfact}. 


\begin{figure*}[t]

	\begin{center}
	    \resizebox{0.7\linewidth}{!}{\input{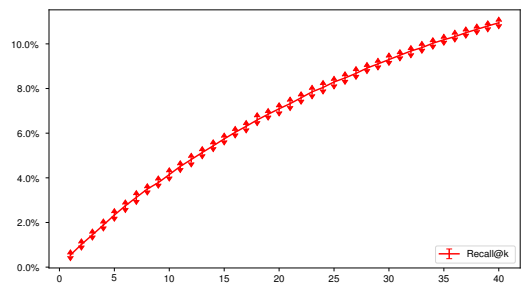}}
	\end{center}
	
	\caption{Domain level recall curve for comparing Politifact's expert sources vs Bing's results. Recall@k is the percentage of domains cited by Politifact that could be retrieved by sampling $k$ random claims from Bing's top 50 results, averaged over all claims in the dataset. Error bars represent 5th and 95th percentiles.}
	\label{fig:ir_recall}
\end{figure*}

\subsection{Source support}
\label{ssec:source-support}

When professional fact checkers are absent, an information retrieval (IR) system is usually used to retrieve potential evidence for a given claim. Yet can IR results always lead to a verdict for the claim, even if inference given sources is perfect?

Consider claim 5 of Table~\ref{tab:taxonomy-examples}. None of the retrieved results are sufficient to judge this claim, yet inferring that this claim is not supported by retrieved results is challenging. It requires reasoning over each source individually as well as across sources.  This is a difficult task even for people to perform. For claim 5 in particular, as we discuss in Subsection~\ref{ssec:solvable-by-humans}, it is unlikely that any set of web sources could be sufficient.

Claim difficulty in this case is evidently linked to the gap of sources that professional fact checkers use as evidence to warrant a verdict, and the search results. Unfortunately, the gap is quite wide, as illustrated in Figure~\ref{fig:ir_recall}. In Table~\ref{tab:pol-vs-bing-sources}, we show that many of the source domains\footnote{We use the term ``source domains'' to refer to the websites that host the retrieved sources, e.g. \textit{www.nytimes.com}.} used in Politifact get low rankings from Bing, so they are likely to be missed if only the top results from search engines are included as evidence sources. For example, many datasets in the literature typically retrieve 5-30 sources \cite{Popat2016CredibilityAO,Baly2018IntegratingSD,multifc,covidfact} while rankings of government pages, social media pages, google docs, etc. that are oft cited by Politifact drop by the hundreds. In practice, this raises the question on the feasibility of simply pairing a verdict given by entities like Politifact with IR results, as in \cite{multifc,fakenewsnet,Popat2016CredibilityAO}, without some consideration of claim difficulty in source support, especially when human judgment on evidence efficacy can be challenging in the first place~\cite{covidfact}.


\begin{table*}[t]
\resizebox{\textwidth}{!}{
\begin{tabular}{@{}p{3.5cm}cp{1.2cm}p{1.2cm}|p{3.5cm}cp{1.2cm}p{1.2cm}@{}}
\toprule
\textbf{Domain}                 & \textbf{\%} & \textbf{Politifact Rank} & \textbf{Bing Rank} & \textbf{Domain}                 & \textbf{\%} & \textbf{Politifact Rank} & \textbf{Bing Rank} \\ \midrule
\textit{www.politifact.com}     & 7.2\%       & 1                        & 2                  & \textit{www.foxnews.com}        & 0.5\%       & 21                       & 43                 \\
\textit{www.nytimes.com}        & 2.5\%       & 2                        & 1                  & \textit{thehill.com}            & 0.4\%       & 22                       & 291                \\
\textit{www.washingtonpost.com} & 2.5\%       & 3                        & 12                 & \textit{www.documentcloud.org}  & 0.4\%       & 23                       & N/A                \\
\textit{www.youtube.com}        & 2.4\%       & 4                        & 11                 & \textit{www.npr.org}            & 0.4\%       & 24                       & 18                 \\
\textit{docs.google.com}        & 1.8\%       & 5                        & 16,752             & \textit{www.snopes.com}         & 0.4\%       & 25                       & 106                \\
\textit{twitter.com}            & 1.6\%       & 6                        & 195                & \textit{www.factcheck.org}      & 0.4\%       & 26                       & 26                 \\
\textit{www.whitehouse.gov}     & 1.3\%       & 7                        & 218                & \textit{www.cbsnews.com}        & 0.4\%       & 27                       & 6                  \\
\textit{www.facebook.com}       & 1.1\%       & 8                        & 189                & \textit{www.usatoday.com}       & 0.4\%       & 28                       & 29                 \\
\textit{www.jsonline.com}       & 1.1\%       & 9                        & 267                & \textit{thomas.loc.gov}         & 0.3\%       & 29                       & N/A                \\
\textit{www.tampabay.com}       & 1.0\%       & 10                       & 1,167              & \textit{www.opensecrets.org}    & 0.3\%       & 30                       & 4,124              \\
\textit{www.politico.com}       & 0.8\%       & 11                       & 31                 & \textit{clerk.house.gov}        & 0.3\%       & 31                       & 33,600             \\
\textit{www.cbo.gov}            & 0.8\%       & 12                       & 1,196              & \textit{www.miamiherald.com}    & 0.3\%       & 32                       & 357                \\
\textit{alt.coxnewsweb.com}     & 0.7\%       & 13                       & N/A                & \textit{www.huffingtonpost.com} & 0.3\%       & 33                       & 36,200             \\
\textit{www.cnn.com}            & 0.7\%       & 14                       & 4                  & \textit{www.kff.org}            & 0.3\%       & 34                       & 224                \\
\textit{www.census.gov}         & 0.7\%       & 15                       & 329                & \textit{www.reuters.com}        & 0.3\%       & 35                       & 39                 \\
\textit{www.senate.gov}         & 0.6\%       & 16                       & 9,367              & \textit{www.nbcnews.com}        & 0.3\%       & 36                       & 300                \\
\textit{www.bls.gov}            & 0.6\%       & 17                       & 475                & \textit{www.statesman.com}      & 0.3\%       & 37                       & 441                \\
\textit{www.cdc.gov}            & 0.6\%       & 18                       & 269                & \textit{www.c-span.org}         & 0.3\%       & 38                       & 621                \\
\textit{www.congress.gov}       & 0.5\%       & 19                       & 8                  & \textit{online.wsj.com}         & 0.3\%       & 39                       & 1,412              \\
\textit{abcnews.go.com}         & 0.5\%       & 20                       & 14                 & \textit{www.forbes.com}         & 0.3\%       & 40                       & 10                 \\ \bottomrule
\end{tabular}
}
\vspace{0.5em}
\caption{ \label{tab:pol-vs-bing-sources} Sources cited by Politifact vs.\ sources retrieved by Bing (at the domain level). Column ``\%'' contains the relative proportion of sources from the corresponding domain being cited by Politifact. Notice that except for \textit{www.politifact.com} and \textit{www.nytimes.com} we see huge discrepencies between the two distributions. For example, Google Docs, the 5th most popular domain in Politifact's citations ranks 16,752 in Bing's citations. Similar trends (though not as extreme) can be seen for social media sites Facebook and Twitter, government websites, etc.}
\end{table*}

\subsection{Source access and readability}
\label{sec:readability}

Assume that a source exists online that provides relevant evidence for fact-checking a given claim, and further assume that it has been indexed by a search engine so that it can be found.  Can it be retrieved by an automated fact-checking system?  If it is locked behind a paywall, such as requiring a source subscription for access, then typically it is useless to automated fact-checking, at least in an academic research context. If it can be retrieved, can it be parsed? Prior work has noted even HTML parsing issues \cite{multifc,Vlachos2014FactCT}, and many cited domains from Politifact span a myriad of other document formats. Table~\ref{tab:source-support} shows that nearly 41.3\% of the sources are hard to parse due to challenges like dynamically rendered javascript, links to pages without the main content (i.e., need more clicks to get to the right content), paywalls, media, etc. We also found a significant portion (7\%) of sources that are tables in various formats, which would require complex reasoning over data~\cite{chen2019tabfact,schlichtkrull2021joint}. Relevant evidence might be buried in a 500 page government report that is too long for existing systems to process well \cite{Wan2019LonglengthLD}, constituting a within-document search problem. 


\subsection{Source trustworthiness}
\label{sec:trustworthiness}

Even if ample evidence sources can be retrieved around a claim, it is important to consider the trustworthiness of these sources. High quality sources such as established news outlets, accredited journalism, scientific research articles, etc.\ are desirable, yet we found that they are unavailable for many claims.
Consider claim 4 of Table~\ref{sec:claim-difficulty-analysis} which talks about the legalese of what constitutes premeditated murder; the best two sources are comments from a transcript of a news show and a blog. The news transcript lands credence to the user comment but is not sufficient by itself to judge the veracity of the claim because it does not specify that the context is ``US law'', which is quite relevant since the claim is about South Africa. In fact, this was a key reason that Politifact ultimately determined that the claim was ``mostly false''.

While such sources can be useful, judging whether to trust them or not is not easy. For example, \citet{Popat2016CredibilityAO} found their per-domain strategy to be extremely limiting. Fact checking systems would likely need to combine information from multiple sources and possibly look at histories (e.g. their past publications and their veracities) of such sources. 
As such, while we are considering source support, it is important to be aware that claims not judgable by verified sources \emph{only} implies another layer of difficulty.

\label{ssec:solvable-by-humans}



\subsection{Inference}
\label{ssec:nli-difficulty}
The last notable challenge we encountered is the difficulty of natural language inference given the evidence. Consider claim 3 of Table~\ref{tab:taxonomy-examples}. The second source, source \textit{b}, directly states that the absence of ICE would not affect border security and refutes the claim. Source \textit{a} is also sufficient by itself to refute the claim as it mentions that CBP does border enforcement while ICE does interior enforcement which allows us to infer that the absence of the latter won't affect border enforcement. But this inference is harder than the first as it requires correctly reasoning about more entities. Another example requiring challenging inference is claim 2 of Table~\ref{tab:taxonomy-examples}. Here, none of the sources jointly talk about ``WI families'' and ``GOP estate tax break'' together and a model would be required to synthesize information from multiple sources. We also saw that Politifact frequently cites government pages containing documents like Senate bills, which contain domain-specific vocabulary often missing from general purpose language models. 

Some claims can be difficult to check even for professional human fact-checkers. For example, consider claim 5 of Table~\ref{tab:taxonomy-examples}. Politifact noted they could not find direct evidence refuting this claim on the web and judged this claim by consulting historians. Indeed, the best retrieved source, source \textit{b}, gets very close to refuting this claim but the phrase ``South of the Union'' is ambiguous. Does it mean states in the south, but inside of the Union, or does it mean states to the south and outside of the union? Since it is unreasonable to expect current systems to be able to check such claims, including such a claim in a dataset does not provide value in benchmarking systems.


\begin{table}[t!]
\centering
\small
\begin{tabular}{@{}llcc@{}}
\toprule
\textbf{} & \textbf{Reason}            & \textbf{\#}  & \textbf{\%}     \\ \midrule
\textbf{Difficult}    & \textbf{}                  & \textbf{165} & \textbf{41.3\%} \\
\midrule
                & \textit{Senate Bills}              & 15           & 3.8\%           \\
                & \textit{Crawl Protections} & 3            & 0.8\%           \\
                & \textit{Generic link}      & 9            & 2.3\%           \\
                & \textit{Indirect link}     & 17           & 4.3\%           \\
                & \textit{Javascript}        & 9            & 2.3\%           \\
                & \textit{Media}             & 15           & 3.8\%           \\
                & \textit{Misc}              & 25           & 6.3\%           \\
                & \textit{Not accessible}    & 21           & 5.3\%           \\
                & \textit{Paywall}           & 11           & 2.8\%           \\
                & \textit{PDFs}              & 12           & 3.0\%           \\
                & \textit{Table}             & 28           & 7.0\%           \\
\midrule
\textbf{OK}     & \textbf{}                  & \textbf{235} & \textbf{58.6\%} \\ \bottomrule
\end{tabular}
\vspace{1em}
\caption{ \label{tab:source-support} Domain level statistics for Politifact's expert sources. We sampled sources from Politifact's top 400 most cited domains and classified domains with respect to parseability.}
\end{table}

\section{Where do we go from here?}

\subsection{Corpora development}
\label{ssec:dataset-design}
We believe analyzing claim difficulty during corpora development can lead to higher-quality data, which directly impact modeling downstream. Dataset development for ``claim+evidence'' based fact checking requires fetching claims, sources and annotations. We discuss opportunities for improvements in these steps separately.

Sources are sometimes retrieved by borrowing expert sources \cite{Vlachos2014FactCT,liarplus,Atanasova2020GeneratingFC}, yet a more realistic setting would be to use search engines~\cite{Popat2016CredibilityAO,Baly2018IntegratingSD,fakenewsnet,multifc}.
However, we call for attention to the wide gap between expert and IR sources, since top IR results may lead to very different sources that may render the claim unjudgable.

Veracity labels are typically crowd sourced \cite{Mitra2015CREDBANKAL,PrezRosas2017AutomaticDO,fever} or taken from fact checking websites \cite{Popat2016CredibilityAO,liar,liarplus,Baly2018IntegratingSD,multifc,Atanasova2020GeneratingFC}.
As practical IR systems are imperfect, claims paired with retrieved sources may be harder to fact check as compared to claims paired with expert sources. If annotations are taken directly from fact checking websites~\cite{Popat2016CredibilityAO,liar,liarplus,Baly2018IntegratingSD,multifc,Atanasova2020GeneratingFC}, the annotations may not be directly portable to the IR evidence sources. 
For such claims, the veracity label should instead be ``not enough info'' (this approach was used by \cite{fever}). 

\subsection{Claim difficulty as a task}
\label{ssec:claim-difficult-as-a-task}
We propose to formally establish claim difficulty prediction as a task, which we argue can be beneficial in multiple fronts.
Firstly, many social media websites and news companies employ human fact checkers, but they can easily be overwhelmed by the vast amount of content being generated today. The output from the claim difficulty prediction task can help triage and prioritize more difficult claims to be checked by human fact-checkers. 
Secondly, claim difficulty prediction can be used to quantify model confidence. Recent works \cite{guo2017calibration,Dong2018ConfidenceMF} have shown that posterior probability with deep models is not a good measure of model performance. The output from the claim difficulty task can provide orthogonal information to quantify model confidence for fact checking models as intuitively they are more likely to make mistakes on difficult claims. 
Lastly, an explainable difficulty prediction model can lead to a more flexible approach for fact checking, for example, to develop model architectures with inductive biases that exploit the specific aspects of the claim difficulty task that we have identified.

\section{Conclusion}

Fact-checking is often a daunting task for human experts \cite{nakov-ijcai2021}, not to mention automated systems. Our analysis shows that some claims are far more difficult to fact-check than others, and we identify a variety of underlying reasons for this: claim ambiguity, source support, source readability, source trustworthiness, and inference challenges. 

Consideration of variable claim difficulty has important conceptual, practical, and research implications for how we pursue fact-checking, including research issues for dataset/task formulation and system architecture. While prior work usefully identified the value of predicting whether or not a given claim is check-worthy \cite{checkthat2020,nakov2021clef}, a claim may be check-worthy yet difficult or impossible to fact-check. By modeling and predicting claim difficulty, we can better prioritize human resources for fact-checking and better distinguish which fact checking problems require deep NLP/IR techniques vs.\ those that might be solved by simpler techniques. It may also be the case sometimes that easier claims are more important to fact-check, e.g., if they are more likely to spread or more likely to cause harm if incorrect yet believed.  Consequently, we view assessment of claim difficulty as one part of a larger assessment in triaging claims so that the fact-checking enterprise can ultimately yield the the greatest societal benefit.

\section{Acknowledgements} 

We thank our anonymous reviewers for their feedback. This research was supported in part by the Knight Foundation, the Micron Foundation, Wipro, and by {\em Good Systems}\footnote{\url{http://goodsystems.utexas.edu/}}, UT Austin's Grand Challenge to develop responsible AI technologies. The statements herein are solely the opinions of the author and not the views of the sponsoring agencies. We acknowledge the Texas Advanced Computing Center (TACC) at The University of Texas at Austin for providing HPC resources that have contributed to the research results reported within this paper.


\bibliographystyle{splncsnat}
\bibliography{bibliography}



\end{document}